%% file: main.tex
\pdfoutput=1

\documentclass[11pt]{article}

\usepackage{acl}

\usepackage{times}
\usepackage{latexsym}

\usepackage{tikz}
\usepackage{pgf}
\usepackage{amsmath}    
\usepackage{amssymb}    
\usepackage{ifthen}
\usepackage{enumitem}
\usepackage{hyperref}       
\usepackage[capitalize,noabbrev]{cleveref}
\usepackage{url}

\definecolor{dgreen}{HTML}{25aa18}

\newcommand{\fixme}[2][]{%
    \ifthenelse{\equal{#1}{}}%
               {\textsl{[\textcolor{red}{#2}]}}%
               {\textsl{[#1: \textcolor{red}{#2}]}}%
    }%

\usetikzlibrary{positioning}
\usetikzlibrary{calc}
\usetikzlibrary{arrows}
\usetikzlibrary{decorations.pathmorphing,decorations.markings}
\usetikzlibrary{shapes}
\usetikzlibrary{shapes.arrows}
\usetikzlibrary{patterns}
\usetikzlibrary{fit}

\usepackage[T1]{fontenc}

\usepackage[utf8]{inputenc}

\usepackage{microtype}

\title{Silo NLP's Participation at WAT2022}

\author{Shantipriya Parida\textsuperscript{\textasteriskcentered}, Subhadarshi Panda\textsuperscript{\dag}, Stig-Arne Grönroos
\textsuperscript{\textasteriskcentered\ddag},\\ \textbf{Mark Granroth-Wilding
}\textsuperscript{\textasteriskcentered}, \textbf{Mika Koistinen
}\textsuperscript{\textasteriskcentered} \\
\textsuperscript{\textasteriskcentered}Silo AI, Helsinki, Finland\\
\texttt{\{firstname.lastname\}@silo.ai}\\
\textsuperscript{\dag}Graduate Center, City University of New York, USA\\
\texttt{spanda@gradcenter.cuny.edu}\\
\textsuperscript{\ddag}University of Helsinki, Finland\\
}

\begin{document}
\maketitle
\begin{abstract}
This paper provides the system description of ``Silo NLP's" submission to the Workshop on Asian Translation (WAT2022). We have participated in the Indic Multimodal tasks (English$\rightarrow$Hindi, English$\rightarrow$Malayalam, and English$\rightarrow$Bengali, Multimodal Translation).
For text-only translation, we trained Transformers from scratch and fine-tuned mBART-50 models.
For multimodal translation, we used the same mBART architecture, and extracted object tags from the images to use as visual features concatenated with the text sequence.

Our submission tops many tasks including English$\rightarrow$Hindi multimodal translation (evaluation test), English$\rightarrow$Malayalam text-only and multimodal translation (evaluation test), English$\rightarrow$Bengali multimodal translation (challenge test), and English$\rightarrow$Bengali text-only translation (evaluation test).    

\end{abstract}

\section{Introduction}
\label{sec:intro}

Machine translation (MT) is a classic sub-field in NLP which investigates the usage of computer software to translate text or speech from one language to another without human involvement \cite{yang2020survey}. Although MT performance has reached near the level of human translators for many high-resource languages, it remains challenging for many low-resource languages 
\cite{popel2020transforming,costa2022no}. Also, effective usage of other modalities (e.g. image) in MT is an important research area in the past few years \cite{sulubacak2020multimodal,parida2021nlphut,parida-etal-2021-multimodal}.

The WAT is an
open evaluation campaign focusing on Asian languages since 2013 \cite{nakazawa2020overview}. In the WAT2022 Multimodal track, a new Indian language \textit{Bengali} was introduced for English$\rightarrow$Bengali text, multimodal translation, and Bengali image captioning task.%
\footnote{\url{https://ufal.mff.cuni.cz/bengali-visual-genome/wat-2022-english-bengali-multim}}

The multimodal translation tasks in WAT2022 consist of image caption translation,
in which the input is a descriptive source language caption together with the image it describes, while the output is a target language caption.
The multimodal input enables the use of image context to disambiguate source words with multiple senses.

In this system description paper, we explain our approach for the tasks (including the sub-tasks) we participated in:
\begin{description}[noitemsep]
    \item [Task~1:] English$\rightarrow$Hindi (EN-HI) Multimodal Translation
    \begin{itemize}[noitemsep]
        \item EN-HI text-only translation
        \item EN-HI multimodal translation
    \end{itemize}
    \item [Task~2:] English$\rightarrow$Malayalam (EN-ML) Multimodal Translation
    \begin{itemize}[noitemsep]
        \item EN-ML text-only translation
        \item EN-ML multimodal translation
    \end{itemize}
    \item [Task~3:] English$\rightarrow$Bengali (EN-BN) Multimodal Translation
    \begin{itemize}[noitemsep]
        \item EN-BN text-only translation
        \item EN-BN multimodal translation
    \end{itemize}
\end{description}

\section{Data sets}
\label{sect_dataset}

We used the data sets specified by the organizer for the related tasks along with additional synthetic data for performance improvement.
The use of additional data places some%
\footnote{All except the EN--HI and EN--ML text-only systems.}
of our submissions in the unconstrained track.

\paragraph{Task 1: English$\rightarrow$Hindi Multimodal Translation} 

For this task, the organizers provided HindiVisualGenome~1.1 \cite{parida2019hindi}\footnote{\url{https://lindat.mff.cuni.cz/repository/xmlui/handle/11234/1-3267}} dataset (HVG for short). The training part consists of 29k English and Hindi short captions of rectangular areas in photos of various scenes and it is complemented by three test sets: development
(D-Test), evaluation (E-Test) and challenge test set (C-Test). Our WAT submissions were for E-Test (denoted ``EV'' in WAT official tables) and C-Test (denoted ``CH'' in WAT tables). 

For the synthetic image features, we use the Flickr8k data set~\citep{hodosh2013framing}.
Even though it is an image captioning data set, we discard the images, treating the data set as in-domain monolingual data.
We use a machine translation into Hindi~\citep{rathi2020deep} as the target side,
and generate image features using the procedure described in Section~\ref{subsec:synthfeats}.

The statistics of the datasets are shown in ~\cref{tab:trainingstat_multimodal_hvg}.

\begin{table}[!htb]
\small
\centering
{%
\begin{tabular}{l@{~~}r@{~~}r@{~~}r@{~~}r@{~~}r}
\hline
 &&\multicolumn{4}{c}{Tokens}  \\
Set & \llap{Sentences} &English  &Hindi & Malayalam & Bengali \\
\hline
Train   & 28930   &143164 &145448 & 107126 & 113978\\
D-Test & 998  	& 4922 &4978 & 3619 & 3936\\
E-Test & 1595   	& 7853 &7852 & 5689& 6408\\
C-Test & 1400   	& 8186 & 8639 & 6044& 6657\\
\hline
\end{tabular}
}
\caption{Statistics of our data used in the  English$\rightarrow$Hindi,  English$\rightarrow$Malayalam, and English$\rightarrow$Bengali Multimodal task: the number of sentences and tokens.
}
\label{tab:trainingstat_multimodal_hvg}
\end{table}

\paragraph{Task 2: English$\rightarrow$Malayalam Multimodal Translation}

For this task, the organizers provided MalayalamVisualGenome~1.0 dataset%
\footnote{\url{https://lindat.mff.cuni.cz/repository/xmlui/handle/11234/1-3533}} (MVG for short). MVG is an extension of the HVG dataset for supporting Malayalam, which belongs to the Dravidian language family \cite{kumar2017morphological}. The dataset size and images are the same as HVG. While HVG contains bilingual English--Hindi segments, MVG contains bilingual English--Malayalam segments, with the English, shared across HVG and MVG, see \cref{tab:trainingstat_multimodal_hvg}.

\paragraph{Task 3: English$\rightarrow$Bengali Multimodal Translation}

For this task, the organizers provided BengaliVisualGenome~1.0 dataset%
\footnote{\url{http://hdl.handle.net/11234/1-3722}} (BVG for short). BVG is an extension of the HVG dataset for supporting Bengali. The dataset size and images are the same as HVG, and MVG, see \cref{tab:trainingstat_multimodal_hvg}.

\section{Experimental Details}
\label{sect_experiment}
This section describes the experimental details of the tasks we participated in.

\subsection{EN-HI, EN-ML, EN-BN text-only translation}

For EN--HI and EN--ML text-only (E-Test and C-Test) translation, we fine-tuned a pre-trained mBART-50 model~\cite{tang2020multilingual}
without using any additional resources. 

For EN-BN text-only (E-Test) translation, we used the Transformer base model as implemented in Open-NMT-Py\footnote{\url{https://opennmt.net/OpenNMT-py/}} using Bangla Natural Language Image to Text (BNLIT) \cite{jishan_bnlit} as an additional dataset. The BNLIT is an image-to-text dataset containing 8743 images and their corresponding text in Bengali. For our experiment, we used Bengali text and translated it into English.  
The training steps are defined for 300K with 5K as validation steps and checkpoint steps.

For C-Test we fine-tuned a pre-trained m-BART-50 model without any additional resources.

\begin{figure}
    \centering
    \includegraphics[width=\linewidth]{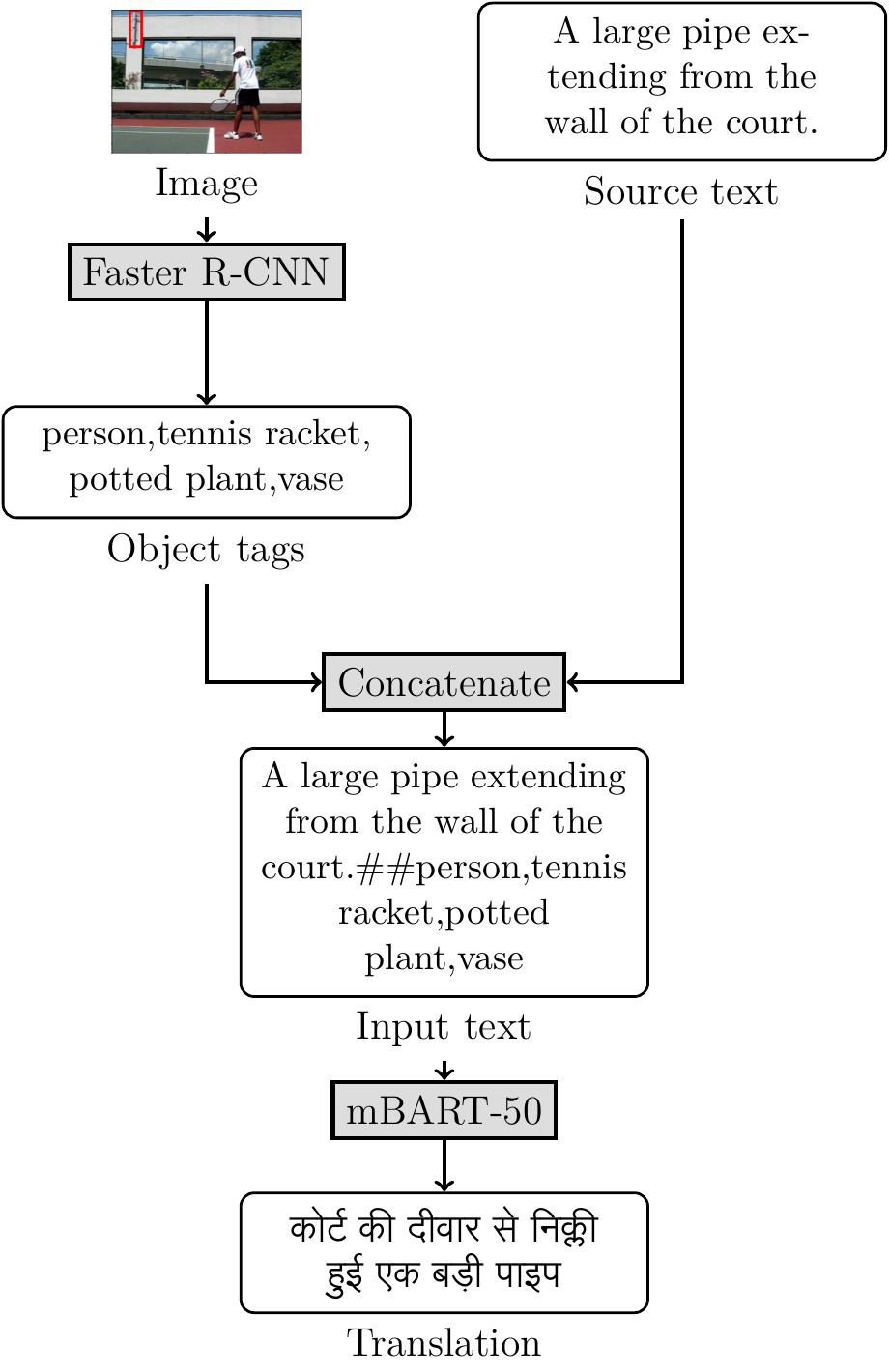}
    \caption{Multimodal translation pipeline.}
    \label{fig:multimodal}
\end{figure}

\subsection{EN-HI, EN-ML, EN-BN Multimodal translation}
Our multimodal translation pipeline is shown in Figure~\ref{fig:multimodal}.
For EN-HI multimodal (E-Test and C-Test) translation, we used the object tags extracted from the HVG dataset images (see ~\Cref{subsect:ext_img_feat}) for image features and concatenated them with the text. Additionally, we used synthetic image features (see ~\cref{subsec:synthfeats}.
The combined data set was used to fine-tune a pre-trained mBART-50 model.

For EN-ML multimodal (E-Test and C-Test) translation, we used object tags extracted from the MVG dataset images and concatenated with the text and fine-tuned on the mBART-50 model.

For EN-BN (E-Test and C-Test) translation, we used object tags extracted from the BVG dataset images and concatenated with the text and fine-tuned on the mBART-50 model.

\subsection{Extracted image features}
\label{subsect:ext_img_feat}

We derive the list of object tags for a given image using the pretrained Faster R-CNN with ResNet-101-C4 backbone. 
It can recognize 80 object types from the COCO data set~\citep{lin2014microsoft}.
Based on their confidence scores, we pick the top 10 object tags. In cases where less than 10 object tags are detected, we consider all the detected tags.
Figure \ref{fig:examples} shows examples of object tags detected for images from the challenge test set.
The detected object tags are then concatenated to the English sentence which needs to be translated to Hindi, Malayalam, and Bengali. The concatenation is done using the special token `\#\#' as the separator. The separator is followed by comma-separated object tags. Adding objects enables the model to utilize visual concepts which may not be readily available in the original sentence.
The English sentences along with the object tags are fed to the encoder of the mBART model.

\input{examples.tex}

\subsection{Synthetic image features}
\label{subsec:synthfeats}

\begin{figure*}
\input{synthfeats.tikz}
\caption{Process for generating synthetic image features. Green arrows indicate processes to synthesize data, and green plus signs indicate the resulting synthetic data. MT is short for machine translation, feat. synth. for image feature synthesis.
\label{fig:synthfeats}}
\end{figure*}

We generate synthetic training data for the multimodal translation task by enriching text-only data using synthetic image features~\citep{gronroos2018memad}.
\citet{gronroos2018memad} use continuous image features and generate the synthetic dummy features by taking the average vector of the features in the training data.
We improve on this procedure by generating discrete features individually for each enriched training example by decoding from a sequence-to-sequence (s2s) model.
The s2s model is trained using the multimodal training data (HVG), but instead of training the normal way, we use both source language and target language text as input (with a separator token between), and our object tags as the output.
The model is a Transformer-base~\citep{vaswani2017attention} model trained using the Marian-NMT~\citep{mariannmt} framework.
The trained model is then used to enrich text-only parallel data with synthetic features.

Our method applies to any text-only parallel data, even though in this experiment we use it to enrich the text from an image captioning data set.
Due to the relatively small size of the multimodal training data sets, domain match of additional training data has great importance for multimodal translation~\citep{gronroos2018memad}.
It is therefore important to apply domain adaptation techniques when using general-domain text-only parallel data.

\section{Results}
\label{sect_result}

We report the official automatic evaluation results of our models for all the participated tasks in \cref{tab:offc_result}.

\begin{table}[!htb]
\begin{center}
\resizebox{\linewidth}{!}{%
\scriptsize
\begin{tabular}{p{3.3cm} rr}
    \hline
      &\multicolumn{2}{c}{WAT BLEU}\\
   System and WAT Task Label & Silo NLP & Best Comp\\ \hline
   \bf{English$\rightarrow$Hindi MM Task} & & \\  \hline
   MMEVTEXT21en-hi& 36.2  & \bf{42.9} \\
   MMEVMM22en-hi & \bf{42}  & 39.4 \\
   MMCHTEXT22en-hi & 29.6 & \bf{41.8} \\
   MMCHMM22en-hi & 39.1   & \bf{39.3} \\
   \hline
   \bf{English$\rightarrow$Malayalam MM Task} & & \\ \hline
   MMEVTEXT21en-ml& \bf{30.8}  & 30.6 \\
   MMEVMM22en-ml & \bf{41}   & -- \\
   MMCHTEXT22en-ml & 14.6      & \bf{19.5} \\
   MMCHMM22en-ml   & \bf{20.4}   & -- \\
   \hline
   \bf{English$\rightarrow$Bengali MM Task} & & \\
   \hline
   MMEVTEXT22en-bn & \bf{41} & \bf{41} \\
   MMEVMM22en-bn   & 42.1  & \bf{43.9} \\
   MMCHTEXT22en-bn & 22.6  & 32.9  \\
   MMCHMM22en-bn   & \bf{28.7} & \bf{28.7} \\
   \hline
\end{tabular}
}
\end{center}
\caption{WAT2022 Automatic Evaluation Results for English$\rightarrow$Hindi, English$\rightarrow$Malayalam, and English$\rightarrow$Bengali. Rows containing ``TEXT" in the task label name denote text-only translation track, and the rows representing ``MM" denote multimodal translation including text and images. 
``--" indicates single submission for the task. For each task, we show the score of our system (Silo NLP) and the score of the best competitor in the respective task. 
}
\label{tab:offc_result}
\end{table}

On the E-Test sets,
our multimodal systems receive the highest BLEU scores for English$\rightarrow$Hindi and English$\rightarrow$Malayalam,
and our text-only systems outperform other text-only systems for English$\rightarrow$Malayalam and English$\rightarrow$Bengali.
Our multimodal systems consistently outperform our text-only systems,
with increases between +1.1 and +10.2 BLEU.

On the C-Test challenge sets,
we have the highest BLEU score for English$\rightarrow$Bengali multimodal translation.
Again, our multimodal systems outperform our text-only systems,
with increases between +5.7 and +9.5 BLEU.

It should be noted that our English$\rightarrow$Hindi multimodal system and English$\rightarrow$Bengali text-only system are unconstrained, making use of substantial additions of in-domain data.
For these language pairs, the increase in translation quality can not be attributed entirely to multimodality.
However, for English$\rightarrow$Malayalam both systems use the same training sentences, making these systems comparable.
The English$\rightarrow$Malayalam multimodal system outperforms the text-only system by +10.2 BLEU for E-test and +5.7 BLEU for C-test.
As the C-test challenge set is constructed to contain translational ambiguity, the improvement is an indication that the image features are useful for disambiguation.

We demonstrate examples of translations obtained for the challenge test set in Figure \ref{fig:examples}.
The extracted image features in the form of object tags are also shown for each image in the figure.
We observe that the multimodal translations are notably better than the text-only translations.
This is consistent with the pattern of the BLEU scores in Table \ref{tab:offc_result}.

\section{Conclusions}
\label{sect_conclusion}
In this system description paper, we presented our system for three tasks in WAT2022: (a) English$\rightarrow$Hindi, (b) English$\rightarrow$Malayalam, and (c) English$\rightarrow$Bengali Multimodal Translation. We released the code through Github for research\footnote{\url{https://github.com/shantipriyap/SiloNLP_WAT2022}}.

In the next step, we will explore the usage of synthetic features in multimodal translation for the Malayalam and Bengali languages, to make use of available text-only corpora for these language pairs.

\section*{Acknowledgements}

We are thankful to Silo AI\footnote{\url{https://silo.ai}}, Helsinki, Finland for the computation resources and the necessary support for participating in WAT2022.

\bibliography{biblio}
\bibliographystyle{acl_natbib}

\end{document}

%% file: examples.tex
\begin{figure*}[t!]
    \centering
    \includegraphics[scale=0.19]{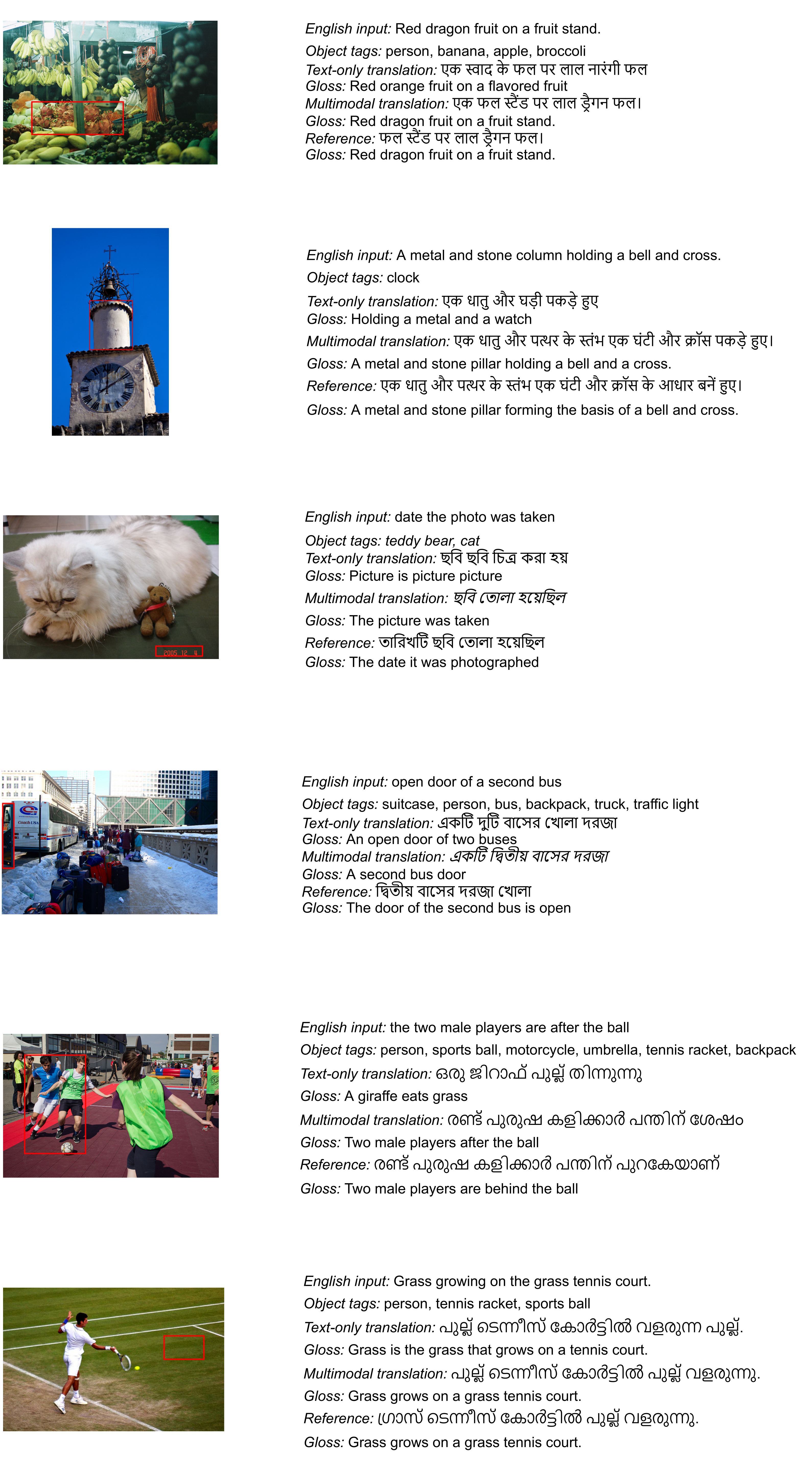}
    \caption{Sample translations for the challenge test set. Both the text-only and multimodal translations are shown. The object tags detected and used in the multimodal translation setup are also shown. Hindi translations are shown for the top two images, Bengali translations are shown for the middle two images, and Malayalam translations are shown for the bottom two images.}
    \label{fig:examples}
\end{figure*}

%% file: synthfeats.tikz.tex
\usetikzlibrary{positioning}
\usetikzlibrary{calc}
\usetikzlibrary{arrows}
\usetikzlibrary{decorations.pathmorphing,decorations.markings}
\usetikzlibrary{decorations.pathreplacing}
\usetikzlibrary{shapes}
\usetikzlibrary{shapes.arrows}
\usetikzlibrary{patterns}
\usetikzlibrary{fit}


\tikzset{ shorten <>/.style={ shorten >=#1, shorten <=#1 } }


\tikzstyle{hilight} = [draw=red, inner sep=0mm, very thick, rounded corners]
\tikzstyle{pil} = [draw, very thick, ->]
\tikzstyle{tabcell} = [text width=30mm, inner sep=0.1mm]
\tikzstyle{heading} = [font=\bf, align=left]
\tikzstyle{textcell} = [align=left]
\tikzstyle{synth} = [color=dgreen, font=\bf]
\tikzstyle{syntharrow} = [draw=dgreen, thick, ->]
\tikzstyle{synthbrace} = [draw=dgreen, thick]

\resizebox{\textwidth}{!}{%
\begin{tikzpicture}

\node[tabcell, heading]                  (h1) {Type of data\strut};
\node[tabcell, heading, right=2mm of h1, text width=40mm] (h2) {Data set\strut};
\node[tabcell, heading, right=2mm of h2] (h3) {Images\strut};
\node[tabcell, heading, right=2mm of h3] (h4) {Source text\strut};
\node[tabcell, heading, right=2mm of h4] (h5) {Target text\strut};
\node[tabcell, heading, right=2mm of h5] (h6) {Sentences\strut};

\node[tabcell, textcell, below=2mm of h1]  (c11) {MMT\strut};
\node[tabcell, textcell, right=2mm of c11, text width=40mm] (c12) {Hindi Visual Genome\strut};
\node[tabcell,           right=2mm of c12] (c13) {\checkmark\strut};
\node[tabcell,           right=2mm of c13] (c14) {\checkmark\strut};
\node[tabcell,           right=2mm of c14] (c15) {\checkmark\strut};
\node[tabcell, textcell, right=2mm of c15] (c16) {29k\strut};

\node[tabcell, textcell, below=2mm of c11] (c31) {Monolingual\strut};
\node[tabcell, textcell, right=2mm of c31, text width=40mm] (c32) {Flickr8k\strut};
\node[tabcell, synth,    right=2mm of c32] (c33) {$\boldsymbol{+}$\strut};
\node[tabcell,           right=2mm of c33] (c34) {\checkmark\strut};
\node[tabcell, synth,    right=2mm of c34] (c35) {$\boldsymbol{+}$\strut};
\node[tabcell, textcell, right=2mm of c35] (c36) {40k\strut};

\draw [decorate,decoration={brace,amplitude=10pt}]
(h3.north west) -- ($(h4.north)+(-2mm,0)$) node [black,midway,yshift=7mm] {Captioning};

\draw [decorate,decoration={brace,amplitude=10pt}]
($(h4.north)+(2mm,0)$) -- ($(h5.north east)+(-6mm,0)$) node [black,midway,yshift=7mm] {Translation};

\draw[syntharrow, out=30, looseness=.5] ($(c34.north west)+(4mm,0)$) to[in=150]
node [synth, midway, yshift=-1mm, above] {MT} (c35.north west);

\draw [synthbrace, decorate,decoration={brace,amplitude=1.5mm}]
($(c35.south west)+(3.5mm,0)$) -- ($(c34.south west)+(0,0)$) node[inner sep=0,midway,yshift=-1.5mm] (midbrace) {};
\draw[syntharrow, out=-140, looseness=.6] (midbrace) to[in=-50]
node [synth, midway, yshift=3mm] {feat. synth.} ($(c33.south west)+(4mm,0)$);

\end{tikzpicture}%
}